\def\year{2020}\relax
\documentclass[letterpaper]{article} 
\usepackage{aaai20}  
\usepackage{times}  
\usepackage{helvet} 
\usepackage{courier}  
\usepackage[hyphens]{url}  
\usepackage{graphicx} 
\usepackage{graphicx}  
\usepackage{amsmath}
\usepackage{amssymb}
\usepackage{xcolor}
\usepackage{subfigure}

\newcommand{\x}{\mathbf{x}}
\newcommand{\ah}{\mathbf{a}}

\newcommand{\z}{\mathbf{z}}
\newcommand{\bh}{\mathbf{b}}
\newcommand{\W}{\mathbf{W}}
\newtheorem{theorem}{Theorem}
\newtheorem{definition}{Definition}
\newtheorem{condition}{Condition}
\newcommand*{\qed}{\hfill\ensuremath{\square}}

\usepackage{multirow}
\usepackage{makecell}
\usepackage{numprint}

\title{Fastened CROWN: Tightened Neural Network Robustness Certificates}

\author{Zhaoyang Lyu,\textsuperscript{\rm 1}\thanks{Equal contribution.
Source code and the appendix are available at https://github.com/ZhaoyangLyu/FROWN.}
\setcounter{footnote}{0}
Ching-Yun Ko,\textsuperscript{\rm 2}\footnote{}
Zhifeng Kong,\textsuperscript{\rm 3}
Ngai Wong,\textsuperscript{\rm 4}
Dahua Lin,\textsuperscript{\rm 1}
Luca Daniel\textsuperscript{\rm 2}\\
\textsuperscript{\rm 1}The Chinese University of Hong Kong, Hong Kong, China\\
\textsuperscript{\rm 2}Massachusetts Institute of Technology, Cambridge, MA 02139, USA\\
\textsuperscript{\rm 3}University of California San Diego, La Jolla, CA 92093, USA\\
\textsuperscript{\rm 4}The University of Hong Kong, Hong Kong, China\\
lyuzhaoyang@link.cuhk.edu.hk, cyko@mit.edu, z4kong@eng.ucsd.edu\\
nwong@eee.hku.hk, dhlin@ie.cuhk.edu.hk, luca@mit.edu
}

\allowdisplaybreaks[4]

\begin{document}

\maketitle

\begin{abstract}
The rapid growth of deep learning applications in real life is accompanied by severe safety concerns.
To mitigate this uneasy phenomenon, much research has been done providing reliable evaluations of the fragility level in different deep neural networks.
Apart from devising adversarial attacks, quantifiers that certify safeguarded regions have also been designed in the past five years.
The summarizing work of~\citeauthor{Salman2019convex} unifies a family of existing verifiers under a convex relaxation framework. We draw inspiration from such work and further demonstrate the optimality of deterministic CROWN~\cite{zhang2018crown} solutions in a given linear programming problem under mild constraints.
Given this theoretical result, the computationally expensive linear programming based method is shown to be unnecessary.
We then propose an optimization-based approach \textit{FROWN} (\textbf{F}astened C\textbf{ROWN}): a general algorithm to tighten robustness certificates for neural networks.
Extensive experiments on various networks trained individually verify the effectiveness of FROWN in safeguarding larger robust regions.
\end{abstract}

\begin{table*}[tbh]
\centering
\caption{List of Notations}
\resizebox{2.1\columnwidth}{!}
{
\begin{tabular}{ll|ll|ll}
\hline
Notation & Definition & Notation & Definition & Notation & Definition \\
\hline
$F: \mathbb{R}^n \to \mathbb{R}^{t}$ & neural network classifier & $\x_0\in\mathbb{R}^n$ & original input & $\x\in\mathbb{R}^n$ & perturbed input \\
$n$ & input size & $\ah^{(k)}$ & the hidden state of the $k$-th layer& $\z^{(k)}$ & the pre-activation of the $k$-th layer\\
$n_k$ & number of neurons in layer $k$ & $[K]$ & set $\{1,2,\cdots,K\}$ & $\mathbb{B}_p(\x_0,\epsilon)$ & $\{\x \mid {\|\x-\x_0\|}_p\leq\epsilon\}$ \\
$F_j^{L}(\x): \mathbb{R}^{n}\to\mathbb{R}$ & linear lower bound of $F_j(\x)$ & $\gamma_j^{(k)L}$ & global lower bound of $\z^{(k)}_j$ & \multirow{2}{*}{$\mathbf{l} \preccurlyeq \z \preccurlyeq \mathbf{u}$} & $\mathbf{l}_r \leq \z_r \leq \mathbf{u}_r$,\\
$F_j^{U}(\x): \mathbb{R}^{n}\to\mathbb{R}$ & linear upper bound of $F_j(\x)$ & $\gamma_j^{(k)U}$ & global upper bound of $\z^{(k)}_j$ & & $\forall\; r\in[s]$, $\mathbf{l}, \z, \mathbf{u}\in\mathbb{R}^s$\\
$s^{[k-1]U}$ & set $\{s^{(1)U},\ldots,s^{(k-1)U}\}$ & $t^{[k-1]U}$ & set $\{t^{(1)U},\ldots,t^{(k-1)U}\}$ & \multirow{2}{*}{$neg(\x)=$} & $\x, \text{if~}\x\leq 0;$\\
$s^{[k-1]L}$ & set $\{s^{(1)L},\ldots,s^{(k-1)L}\}$ & $t^{[k-1]L}$ & set $\{t^{(1)L},\ldots,t^{(k-1)L}\}$ && $0, \text{otherwise}.$\\
$\ah^{[k]}$ & set $\{\ah^{(1)}, ,\ah^{(2)},...,\ah^{(k)}\}$ & $\z^{[k]}$ & set $\{\z^{(1)}, ,\z^{(2)},...,\z^{(k)}\}$ & $\sigma$ & ReLU/ Sigmoid/ Tanh activation\\
\hline 
\end{tabular}
}
\label{tbl:notation}
\end{table*}

\section{Introduction}
\label{sec:intro}
The vulnerability of deep neural networks remains an unrevealed snare in the beginning years of the deep learning resurgence.
In~\citeyear{szegedy2014intriguing},~\citeauthor{szegedy2014intriguing} uncovered the discovery of hardly-perceptible adversarial perturbations that could fool image classifiers.
This discovery agonized the fast development of accuracy-oriented deep learning and shifted community's attentions to the fragility of trained models.
Especially with the increasing adoption of machine learning and artificial intelligence in safety-critical applications, the vulnerability of machine learning models to adversarial attacks has become a vital issue~\cite{sharif2016accessorize,kurakin2017adversarial,carlini2017Towards,wong2018provable}.
Addressing this urging issue requires reliable ways to evaluate the robustness of a neural network, namely by studying the safety region around a data point where no adversarial example exists.
This understanding of machine learning models' vulnerability will, on the other hand, help industries build more robust intelligent systems.

Disparate ways of reasoning and quantifying vulnerability (or robustness) of neural networks have been exploited to approach this dilemma, among which \textit{attack-based} methods have long been in a dominating position. 
In these years, a sequel of adversarial attack algorithms have been proposed to mislead networks' predictions in tasks such as object detection~\cite{goodfellow2015explaining,moosavi2016deepfool}, 
visual question answering~\cite{mudrakarta2018did,Zeng2019CVPR,Gao2019ICCV,Gao2019CVPR},
text classification~\cite{papernot2016crafting}, 
speech recognition ~\cite{cisse2017houdini,Yuan2017Crafting}, and audio systems~\cite{Carlini2018Audio}, where the level of model vulnerability is quantified by the distortion between successful adversaries and the original data points.
Notably, the magnitudes of distortions suggested by attack-based methods are essentially upper bounds of the minimum adversarial distortion. 

In contrast to attack-based approaches, attack-agnostic \textit{verification-based} methods evaluate the level of network vulnerability by either directly estimating~\cite{szegedy2014intriguing,weng2018towards} or lower bounding~\cite{hein2017formal,raghunathan2018certified,dvijotham2018dual,zhang2018crown,Singh2018Fast,weng2019proven} the minimum distortion networks can bear for a specific input sample. As an iconic robustness estimation, CLEVER~\cite{weng2018evaluating} converts the robustness evaluation task into the estimation of the local Lipschitz constant, which essentially associates with the maximum norm of the local gradients \textit{w.r.t.} the original example.
Extensions of CLEVER~\cite{weng2018extensions} focuses on twice differentiable classifiers and works on first-order Taylor polynomial with Lagrange remainder. 

A number of verification-based methods have been proposed in literature to compute a lower bound of the safeguarded region around a given input, \textit{i.e.} a region where the network is guaranteed to make consistent predictions despite any input perturbations.
A pioneering work in providing certifiable robustness verification~\cite{szegedy2014intriguing} computes the product of weight matrix operator norms in ReLU networks to give a lower-bounding metric of the minimum distortion.
However, this certificate method was shown to be generally too conservative to be useful~\cite{hein2017formal,weng2018towards}.
Later, tighter bounds have also been provided for continuously-differentiable shallow networks by utilizing local Lipschitz constants of the network~\cite{hein2017formal}.
Then, for the first time, the formal verification problem is reduced from a mixed integer linear programming (MILP) problem to a linear programming (LP) problem when dealing with $l_\infty$-norm box constraints~\cite{wong2018provable}.
Its concurrent works include
Fast-Lin~\cite{weng2018towards}, which analytically calculates bounds for perturbed samples in given regions and finds the largest certifiable region for ReLU networks through binary search.
Fast-Lin is further generalized for multilayer perceptrons with general activation functions in CROWN~\cite{zhang2018crown}.
Recently,~\citeauthor{Salman2019convex} conclude a general framework for a genre of convex relaxed optimization problems and demonstrate existing approaches to be special cases of their proposal. 
Notably, although~\citeauthor{wong2018provable} propose to verify the robustness for ReLU network by the use of LP, 
a feasible dual solution is instead used in practice to avoid any actual use of LP solvers.
Comparatively,~\citeauthor{Salman2019convex} experiment with more than one linear function to bound nonlinear activations (\textit{e.g.} $2$ lower-bounding functions for ReLU act.) and stick to LP solvers.

Certifiable robustness lower bounds are especially vital in safety-critical scenarios (\textit{e.g.} autonomous driving car) since any miss-classification can be lethal.
However, albeit being useful, new challenges with these certifiable quantifiers arise. There are, in most cases, non-negligible gaps between the certified lower and upper bounds of the minimum distortion. 
This inconsistency in the quantification questions diminishes the use of these state-of-the-art robustness evaluation approaches.

In this article, we stay in line with the previous sequel of works that focus on linear bounds and provide two major contributions:
\begin{enumerate}
    \item We prove that if we limit the constraint relaxation to be exactly one linear bound in each direction (upper and lower) in the LP-based method, the results provided by CROWN are optimal. Therefore the costly LP solving process is unnecessary under this relaxation.
    \item We propose a general optimization framework that we name \textit{FROWN} (\textbf{F}astened C\textbf{ROWN}) for tightening the certifiable regions guaranteed by CROWN, which is also theoretically applicable to tighten convolutional neural network certificate CNN-Cert~\cite{Akhilan2019CNN-Cert} and recurrent neural network certificate POPQORN~\cite{ko2019quantifying}.
\end{enumerate}

\section{Backgrounds} 
\label{sec:background}
This section summarizes the most relevant backgrounds of our proposals. Specifically, LP formulation~\cite{Salman2019convex} is summarized, together with the seminal work of Fast-Lin~\cite{weng2018towards} and CROWN~\cite{zhang2018crown} (generalized Fast-Lin). We first begin by giving the definition of an $m$-layer neural network:
\paragraph{Definitions.} 
Given a trained $m$-layer perceptron $F$, we denote the hidden unit, weight matrix, bias, and pre-activation unit of the $k$-th layer ($k\in[m]$) as $\ah^{(k)}$, $\W^{(k)}$, $\bh^{(k)}$, and $\z^{(k)}$, respectively. Hence, ${\z}^{(k)} = {\W}^{(k)} {\ah}^{(k-1)} + {\bh}^{(k)}, ~{\ah}^{(k)} = \sigma({\z}^{(k)})$,
where $\ah^0 = \x_0 \in \mathbb{R}^n$ is the original input and $F(\x) = \z^{(m)}$ is the network output. Denoting the number of neurons as $n_k$ for the $k$-th layer, implies that $\ah^{(k)}, \z^{(k)}, \bh^{(k)} \in \mathbb{R}^{n_k}$ and $\W^{(k)} \in \mathbb{R}^{n_k \times n_{k-1}}$, for $k\in[m]$. 
Furthermore,  we use square brackets in the superscripts to group a set of variables (\textit{e.g.} $\ah^{[m]}$ denotes the set of variables $\{\ah^{(1)}, ,\ah^{(2)},...,\ah^{(m)}\}$ and $\z^{[m]}$ denotes the set of variables $\{\z^{(1)}, ,\z^{(2)},...,\z^{(m)}\}$). Table~\ref{tbl:notation} summarizes all the notations we use in this paper. 

When quantifying the robustness of the $m$-layer neural network, one essentially wants to know 1) how far the network output will deviate when the input is perturbed with distortions of a certain magnitude and 2) the critical point in terms of distortion magnitudes, beyond which the deviation might alter the model prediction.
If we let $\x\in\mathbb{R}^n$ denote the perturbed input of $\mathbf{x_0}$ (class $i$) within an $\epsilon$-bounded $l_p$-ball (\textit{i.e.}, $\x\in\mathbb{B}_p(\x_0,\epsilon), \text{or} {\|\x-\x_0\|}_p\leq\epsilon$), the task of robustness analysis for this network intrinsically involves the comparison between the $i$-th network output $F_i(\x)$ and other outputs $F_{j\neq i}(\x)$. 
In practice, one can translate the original problem to the problem of deriving a lower bound of $F_i(\x)$ and upper bounds of $F_{j\neq i}(\x)$ for perturbed inputs within the $l_p$-norm ball.
With such quantifier, network $F$ is guaranteed to make consistent predictions within the $l_p$-norm ball if the lower bound of the original class output is always larger than the upper bounds of all other classes' outputs.

We summarize below the LP problem~\cite{Salman2019convex} to solve the lower bound of $F_i(\x)$. The LP problem for the upper bound of $F_{j\neq i}(\x)$ can be similarly derived.

\paragraph{The LP problem.} 
The optimization problem for solving the lower bound of $F_i(\x) = \z^{(m)}_i = {\W^{(m)}_{i,:} \ah^{(m-1)} + \bh^{(m)}_i}$ reads:
\begin{align}
\min_{\ah^{(0)} \in \mathbb{B}_p(x_0, \epsilon),\ah^{[m-1]},\z^{[m-1]}}  ~{{\W^{(m)}_{i,:} \ah^{(m-1)} + \bh^{(m)}_i}}\label{eqn:nonlinearconstraint} \\
\nonumber\text{s.t.~}
\begin{cases}
    \z^{(k)} &= \W^{(k)} \ah^{(k-1)} + \bh^{(k)}, \forall~ k\in[m-1],\\
    \ah^{(k)} &= \sigma(\z^{(k)}), \forall~ k\in[m-1].
\end{cases}
\end{align}
The optimization problem for upper bounds can be readily obtained by replacing the ``min'' operation by ``max''. Then, the nonlinear constraint in~\eqref{eqn:nonlinearconstraint} is lifted with linear relaxations. Specifically, suppose the lower and upper bounds of the pre-activation units $\z^{[m-1]}$ are known, namely, for $k$ from $1$ to $m-1$, as a result $\mathbf{l}^{(k)}$ and $\mathbf{u}^{(k)}$ satisfy
\begin{equation}
    \label{ieqn:bound_preactivation}
    \mathbf{l}^{(k)} \preccurlyeq \z^{(k)} \preccurlyeq \mathbf{u}^{(k)},
\end{equation}
and therefore every element $\sigma(\z^{(k)}_i)$ of the nonlinear activation $\sigma(\z^{(k)})$ in constraint~\eqref{eqn:nonlinearconstraint} can be bounded by linear functions:
\begin{align}
\label{ieqn:linearfuncs}
    h^{(k)L}_i(\z^{(k)}_i) \leq \sigma(\z^{(k)}_i) \leq h^{(k)U}_i(\z^{(k)}_i), \forall \z^{(k)}_i \in [\mathbf{l}^{(k)}_i, \mathbf{u}^{(k)}_i],
\end{align}
for $i \in [n_k]$.
The existence of linear bounding functions in~\eqref{ieqn:linearfuncs} is guaranteed since $\z^{(k)}_i$ is bounded and compactness is a continuous invariant within any interval. For example, the following are bounding functions: $h^{(k)L}_i(\z^{(k)}_i)=\min_{z^{(k)}_i\in[\mathbf{l}^{(k)}_i,\mathbf{u}^{(k)}_i]}(\sigma(\z^{(k)}_i)), h^{(k)U}_i(\z^{(k)}_i)=\max_{\z^{(k)}_i\in[\mathbf{l}^{(k)}_i,\mathbf{u}^{(k)}_i]}(\sigma(\z^{(k)}_i))$.
$h^{(k)L}_i$ and $h^{(k)U}_i$ can also be taken as the pointwise supremum and infimum of several linear functions, respectively, which is equivalent to using multiple linear constraints.
In practice, \citeauthor{Salman2019convex} use linear functions characterized by slopes and intercepts:
\begin{align}
    \nonumber h^{(k)L}_i(\z^{(k)}_i) = s^{(k)L}_i \z^{(k)}_i + t^{(k)L}_i,\\
    h^{(k)U}_i(\z^{(k)}_i) = s^{(k)U}_i \z^{(k)}_i + t^{(k)U}_i.\label{eqn:linearfuncs}
\end{align}
The optimization problem can therefore be relaxed to an LP-alike problem\footnote{The optimization problem turns to a strict LP problem only when we have $p = \infty$ or $1$ that makes the feasible set a polyhedron. However we coarsely denote all the cases in general as LP problems since all the constraints are now linear in the variables.}:
\begin{align}
&\min_{\ah^{(0)} \in \mathbb{B}_p(x_0, \epsilon),\ah^{[m-1]},\z^{[m-1]}}  ~{{\W^{(m)}_{i,:} \ah^{(m-1)} + \bh^{(m)}_i}} \label{eqn:LP}\\
\nonumber~~&\text{s.t.~}
\begin{cases}
&\hspace{-1em}\z^{(k)} = \W^{(k)} \ah^{(k-1)} + \bh^{(k)},\forall~ k\in[m-1],\\
&\hspace{-1em}h^{(k)L}(\z^{(k)})  \preccurlyeq \ah^{(k)} \preccurlyeq h^{(k)U}(\z^{(k)}), \forall~ k\in[m-1],\\
&\hspace{-1em}\mathbf{l}^{(k)} \preccurlyeq \z^{k} \preccurlyeq \mathbf{u}^{(k)}, \forall~ k\in[m-1].\\
\end{cases}
\end{align}

Recalling that with the optimization formed as in Problem~\eqref{eqn:LP}, one is essentially optimizing for the lower (or upper) output bounds of the network (the pre-activation of the $m$-th layer), whereas these build upon the assumption that the pre-activation bounds are known as Equation~\eqref{ieqn:bound_preactivation}. To satisfy this assumption, one actually only needs to substitute the layer index $m$ in Problem~\eqref{eqn:LP} with the corresponding intermediate layer's index. In practice, one can recursively solve LP problems from the second layer to the $m$-th layer to obtain the pre-activation bounds for all layers. In this process, the pre-activation bounds computed in a layer also constitute the optimization constraint for the next to-be-solved optimization problem for the next layer's pre-activation. 
See details of this LP-based method in Appendix Section A.3.

\paragraph{CROWN Solutions.}
Here we briefly walk through the derivation of Fast-Lin~\cite{weng2018towards} and CROWN~\cite{zhang2018crown}, whose procedures are essentially the same except for activation-specific bounding rules adopted. The first steps include bounding $\z^{(k)}_i$, $k\in[m]$
\footnote{Similar to the discussion in the LP-based method above, the bounds computed are exactly network output bounds when $k=m$; whereas $k\neq m$ gives the pre-activation bounds to fulfill the assumption in Inequality~\eqref{ieqn:bound_preactivation}.}.
\begingroup
\small
\begin{align}
    {\z}^{(k)}_i &= \sum^{n_{k-1}}_{j=1}{\W}^{(k)}_{i,j} \sigma({\z}^{(k-1)}_j) + {\bh}^{(k)}_i, \label{eqn:crown1}\\
    &\geq \sum^{n_{k-2}}_{j=1}{\tilde{\W}}^{(k-1)}_{i,j} \sigma({\z}^{(k-2)}_j) + {\tilde{\bh}}^{(k-1)}_i, \label{eqn:crown2}
\end{align}
\endgroup
where $neg(\x)=\x$, if $\x\leq 0$; $neg(\x)=0$, otherwise. And 
${\tilde{\W}}^{(k-1)}_{i,j} = [relu({\W}^{(k)}_{i,:})\odot (s^{(k-1)L})^{\top}+ neg({\W}^{(k)}_{i,:})\odot (s^{(k-1)U})^{\top}] \W^{(k-1)}_{:,j},~{\tilde{\bh}}^{(k-1)}_i = [relu({\W}^{(k)}_{i,:})\odot (s^{(k-1)L})^{\top}+ neg({\W}^{(k)}_{i,:})\odot (s^{(k-1)U})^{\top}] {\bh}^{(k-1)}+ relu({\W}^{(k)}_{i,:}) t^{(k-1)L}+ neg({\W}^{(k)}_{i,:}) t^{(k-1)U} + {\bh}^{(k)}_i,$
where $\odot$ denotes element-wise products.
As Equations~\eqref{eqn:crown1} and~\eqref{eqn:crown2} are in similar forms, the above procedures can be repeated until all the nonlinearities in $k-1$ layers are unwrapped by linear bounding functions and ${\z}^{(k)}_i$ is upper bounded by $\sum^{n}_{j=1}{\tilde{\W}}^{(1)}_{i,j} \x_j + {\tilde{\bh}}^{(1)}_i$, where ${\tilde{\W}}^{(1)}_{i,j}$ and ${\tilde{\bh}}^{(1)}_i$ are similarly defined as shown above. Taking the dual form of the bound then yields the closed-form bound $\gamma_j^L$ that satisfies
\begin{align}
    {\z}^{(k)}_i \geq \gamma_i^{(k)L} :=& {\tilde{\W}}^{(1)}_{i,:} \x_0 - \epsilon\|{\tilde{\W}}^{(1)}_{i,:}\|_q + {\tilde{\bh}}^{(1)}_i,\label{eqn:crown}
\end{align}
$\forall~ \x \in \mathbb{B}_p(\x_0, \epsilon)$, where $1/p+1/q=1$. Although the steps above are used to derive the closed-form lower bound, the closed-form upper bound $\gamma_i^{(k)U}$ can be similarly derived. 
To quantify the robustness for an $m$-layer neural network, one needs to recursively adopt formulas in Equation~\eqref{eqn:crown} to calculate the bounds of pre-activation\footnote{$\z^{(1)}$ is deterministically computed by the input and $\z^{(m)} = F(\x)$ is the output bound.} $\z^{(k)}$, for $k = 2,\ldots,m$. These bounds, as will be explained in more details later, confine the feasible set for choosing bounding linear functions in Equation~\eqref{eqn:linearfuncs}.
Notably, lower bound $\gamma_i^{(k)L}$ and upper bound $\gamma_i^{(k)U}$ are implicitly reliant to slopes of bounding lines in previous layers $s^{[k-1]U} = \{s^{(1)U},\ldots,s^{(k-1)U}\}$, $s^{[k-1]L} = \{s^{(1)L},\ldots,s^{(k-1)L}\}$ and their intercepts $t^{[k-1]U} = \{t^{(1)U},\ldots,t^{(k-1)U}\}$, $t^{[k-1]L} = \{t^{(1)L},\ldots,t^{(k-1)L}\}$. 
A major difference that distinguishes CROWN from our contributions in the following sections is its deterministic rules of choosing upper/lower-bounding lines. The readers are referred to the literature~\cite{zhang2018crown} or Sections~A.2 and A.4 in the appendix for more details of CROWN.

\section{Relation Between the LP Problem and CROWN Solutions}
\label{sec:crown_equ_lp}
Now we discuss the relationship between the LP problem formulation and CROWN. A key conclusion is that: CROWN is not only a dual feasible solution of the presented LP problem as discussed by~\citeauthor{Salman2019convex}, it gives the optimal solution under mild constraints. 

Before introducing the optimality of CROWN solutions under the LP framework, we define an important condition in the computation process of CROWN as below:
\begin{condition}
\label{cdt:self_consistency}
\textbf{Self-consistency.} Suppose $\{\tilde{s}^{[v-1]U}, \tilde{s}^{[v-1]L},$ $\tilde{t}^{[v-1]U}, \tilde{t}^{[v-1]L}\}$ are used to calculate $\gamma_i^{(v)L}$ and $\gamma_i^{(v)U}$, $\{\hat{s}^{[k-1]U},\hat{s}^{[k-1]L},\hat{t}^{[k-1]U},\hat{t}^{[k-1]L}\}$ are used to calculate $\gamma_j^{(k)L}$ and $\gamma_j^{(k)U}$, then the following condition holds,
\begin{align*}
    \tilde{s}^{[v-1]U} &= \hat{s}^{[v-1]U}, \tilde{s}^{[v-1]L} = \hat{s}^{[v-1]L},\\
    \tilde{t}^{[v-1]U} &=
    \hat{t}^{[v-1]U},\ \tilde{t}^{[v-1]L} = \hat{t}^{[v-1]L},
\end{align*}
for $\forall~ i \in [n_v], \forall j \in [n_k], 2\leq v \leq k\leq m$ and two sets equal to each other is defined as their corresponding elements equal to each other.
\end{condition}
A similar condition can also be defined in the LP-based method and is supplemented in Section~A.3 in the appendix.
The self-consistency condition guarantees the same set of bounding lines is used when computing bounds for different neurons in the process of CROWN or the LP-based method. We note that both the original CROWN and the LP-based method satisfy the self-consistency condition.
\begin{theorem}
\label{thm:crownoptimal}
The lower bound obtained by Equation~\eqref{eqn:crown} is the optimal solution to Problem~\eqref{eqn:LP} when the following three conditions are met:
\begin{itemize}
    \item Each of the $h^{(k)L}(\z^k)$ and $h^{(k)U}(\z^k)$ in Problem~\eqref{eqn:LP} is chosen to be \underline{one} linear function\footnote{Theoretically one can use multiple linear functions to bound the nonlinearity in Problem~\eqref{eqn:LP} to obtain tighter bounds.} as in Equation~\eqref{eqn:linearfuncs}.
    \item The LP problem shares the same bounding lines with CROWN.
    \item The self-consistency conditions for both CROWN and the LP-based method hold. 
\end{itemize}
\end{theorem}
We refer readers to Section~A.5 in the appendix for the proof. 
We emphasize the cruciality of the self-consistency conditions in Theorem~\ref{thm:crownoptimal}: We do observe CROWN and the LP-based method can give different bounds when Condition~\ref{cdt:self_consistency} is not met, even though the two use the same bounding lines. In essence, Theorem~\ref{thm:crownoptimal} allows one to compute bounds analytically and efficiently following
steps in CROWN instead of solving the expensive LP problems under certain conditions.

\begin{figure}[t]
    \centering
    \includegraphics[width=0.8\columnwidth]{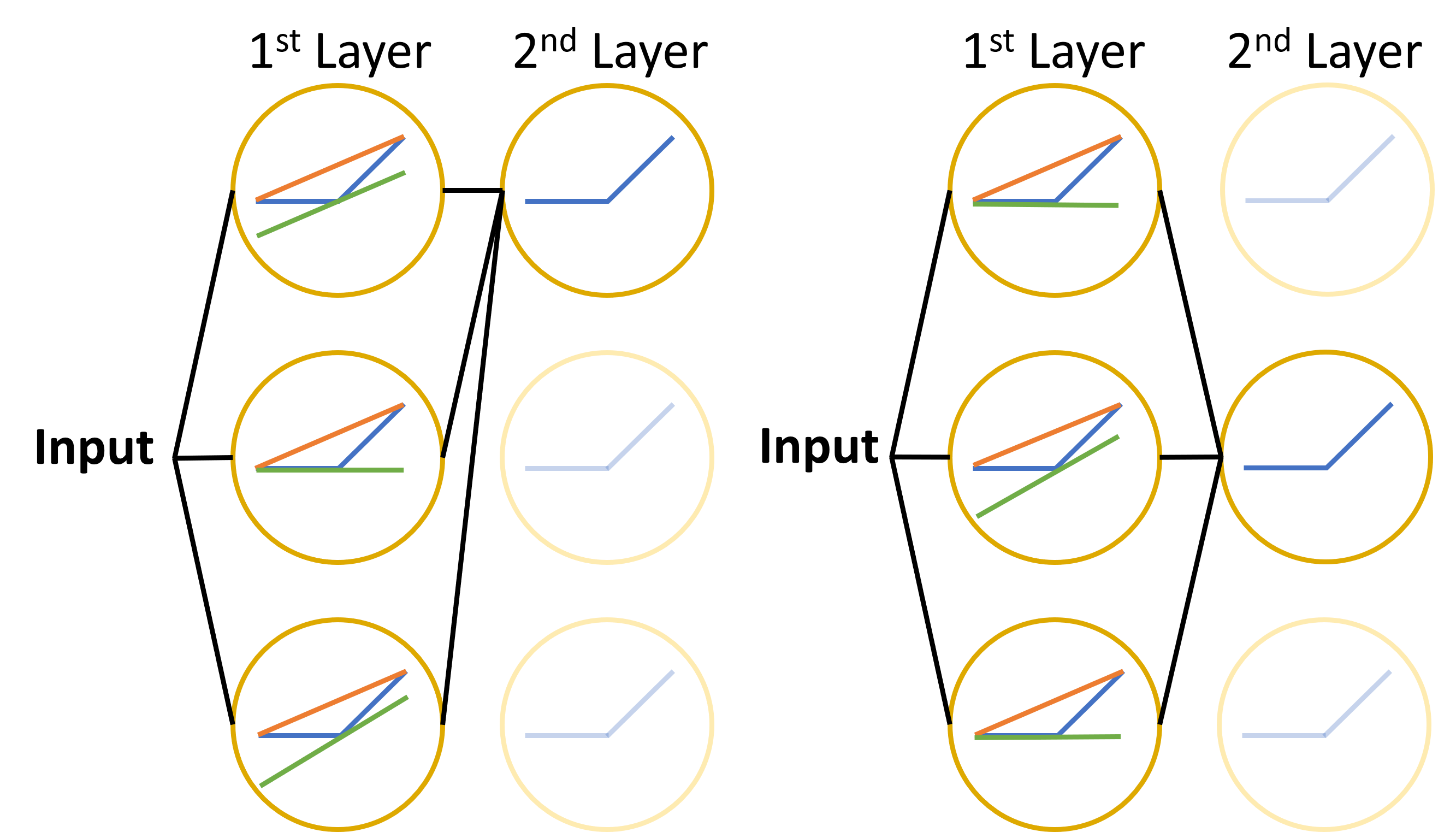}
    \caption{The process of CROWN using different bounding lines to compute the closed-form bounds for different neurons. The blue curves are the ReLU activation. The orange and green lines are the upper and lower bounding lines, respectively. When computing closed-form bounds of the pre-activation of neurons in the second layer, different neurons can choose different bounding lines in the previous layers to yield the tightest closed-form bounds for themselves.}
    \label{fig:neuronwise_optmize}
\end{figure}

\begin{table*}
\centering
\caption{Search space of bounding lines for ReLU, Sigmoid, and Tanh  functions. 
``Variable" is the optimization variable that characterizes
the bounding line. ``Range" is the feasible region of the variable. 
``-" indicates the case when the tightest bounding line is unique and chosen. 
The slope and intercept of ReLU upper-bounding line are always set to be $s_0$ and $t(s_0, l)$, respectively. 
 } 
\label{tbl:bounding_line_search_space}
\resizebox{2.1\columnwidth}{!}{
\begin{tabular}{c|ccc|cccc|cccc}
\noalign{\hrule height 0.75pt}
Nonlinearity & \multicolumn{3}{c|}{ReLU (Lower bnd.)} & \multicolumn{4}{c|}{Sigmoid \& Tanh (Upper bnd.)} & \multicolumn{4}{c}{Sigmoid \& Tanh (Lower bnd.)}\\\hline
Pre-activation   & \multirow{2}{*}{$l<u \leq 0$}   & \multirow{2}{*}{$l<0 < u$} & \multirow{2}{*}{$0 \leq l<u$}  & \multirow{2}{*}{$l<u \leq 0$}   & \multicolumn{2}{c}{$l<0<u$}                & \multirow{2}{*}{$0 \leq l<u$} & \multirow{2}{*}{$l<u \leq 0$} & \multicolumn{2}{c}{$l<0<u$} & \multirow{2}{*}{$0 \leq l<u$} \\
Bounds & & & & & case 1 & case 2 & & & case 3 & case 4 &\\\hline
Variable & - & $s$  & - & - & $d_1$ & - & $d_1$ & $d_2$ & $d_2$ &  - & - \\
Range & - & $[0,1]$ & - & - & $[l_d, u]$ & - & $[l, u]$ & $[l, u]$ & $[l, u_d]$ & - & - \\
Slope & $s_0$ & $s$ & $s_0$ & $s_0$ & $\sigma'(d_1)$ & $s_0$ & $\sigma'(d_1)$ & $\sigma'(d_2)$ & $\sigma'(d_2)$  & $s_0$ & $s_0$ \\
Intercept & $t(s_0,l)$ & $0$ & $t(s_0,l)$  & $t(s_0, l)$ & $t(\sigma'(d_1), d_1)$ & $t(s_0, l)$ & $t(\sigma'(d_1), d_1)$ & $t(\sigma'(d_2), d_2)$ & $t(\sigma'(d_2), d_2)$ & $t(s_0, l)$ & $t(s_0, l)$ \\\hline
\multicolumn{12}{l}{Notes: Case 1 refers to $\sigma'(u) l + t(\sigma'(u),u) \geq \sigma(l)$; and case 2, otherwise. Case 3 refers to $\sigma'(l) u + t(\sigma'(l),l) \leq \sigma(u)$; and case 4, otherwise. $s_0 = [\sigma(u) -\sigma(l)]/$}\\
\multicolumn{12}{l}{$(u-l)$, $t(s,y)=\sigma(y)-s y$.
$l_d$ and $u_d$ are defined as the abscissas of the points at which the tangent passes the left endpoint $(l, \sigma(l))$ and the right endpoint $(u, \sigma(u))$,} \\
\multicolumn{12}{l}{ respectively.
$d_1$ and $d_2$ are the abscissas of the points of tangency.
See Figure~\ref{fig:sigmoid} for the visualization of $l_d$, $u_d$, $d_1$ and $d_2$.
}\\
\noalign{\hrule height 0.75pt}     
\end{tabular}
}
\end{table*}

\begin{figure}[tbh]
    \centering
    \subfigure[$0 \leq l < u$]{
    \label{fig:sigmoid1}
    \includegraphics[width=0.49\columnwidth]{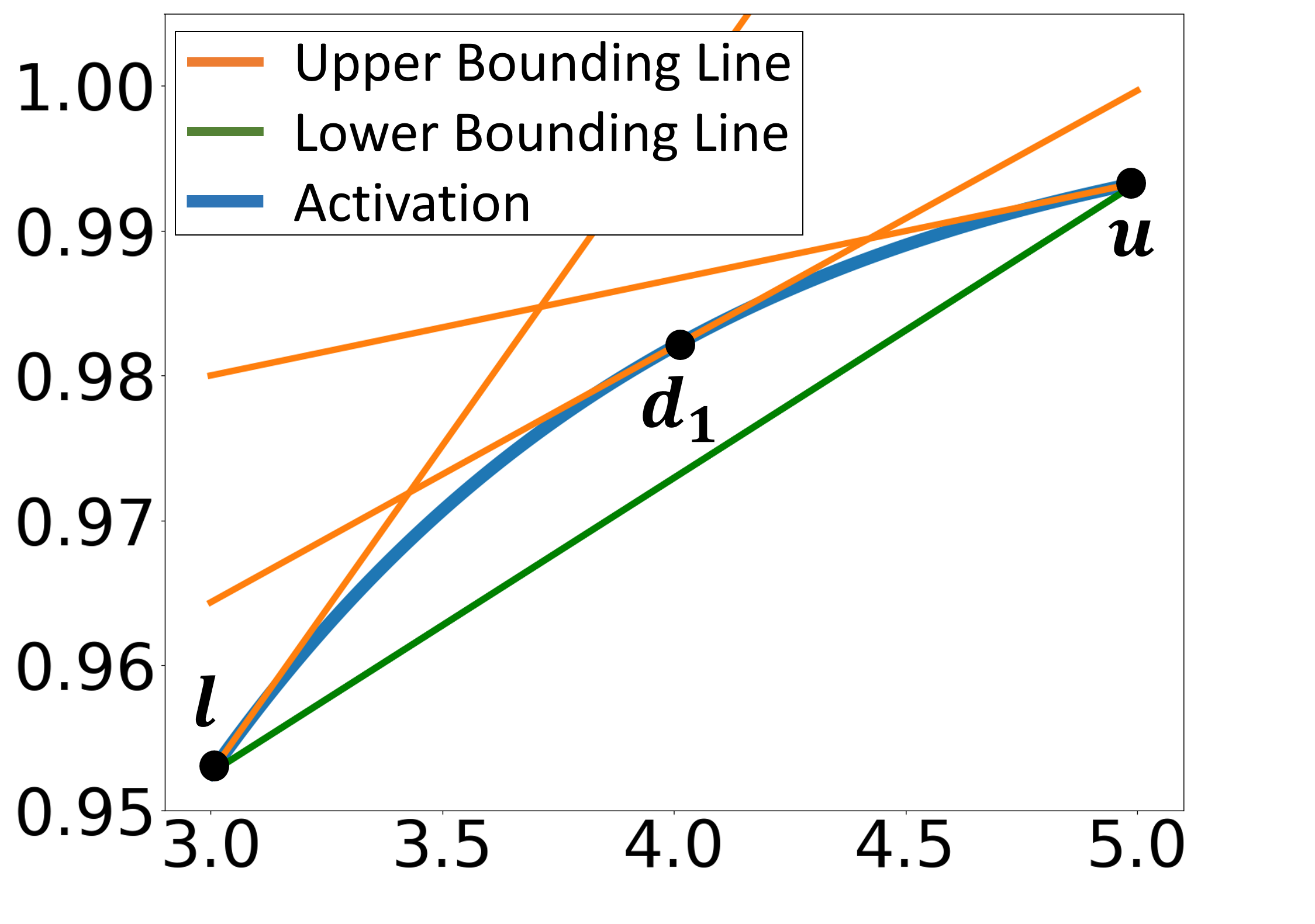}}
    \subfigure[$l < 0 < u$]{
    \label{fig:sigmoid2}
    \includegraphics[width=0.49\columnwidth]{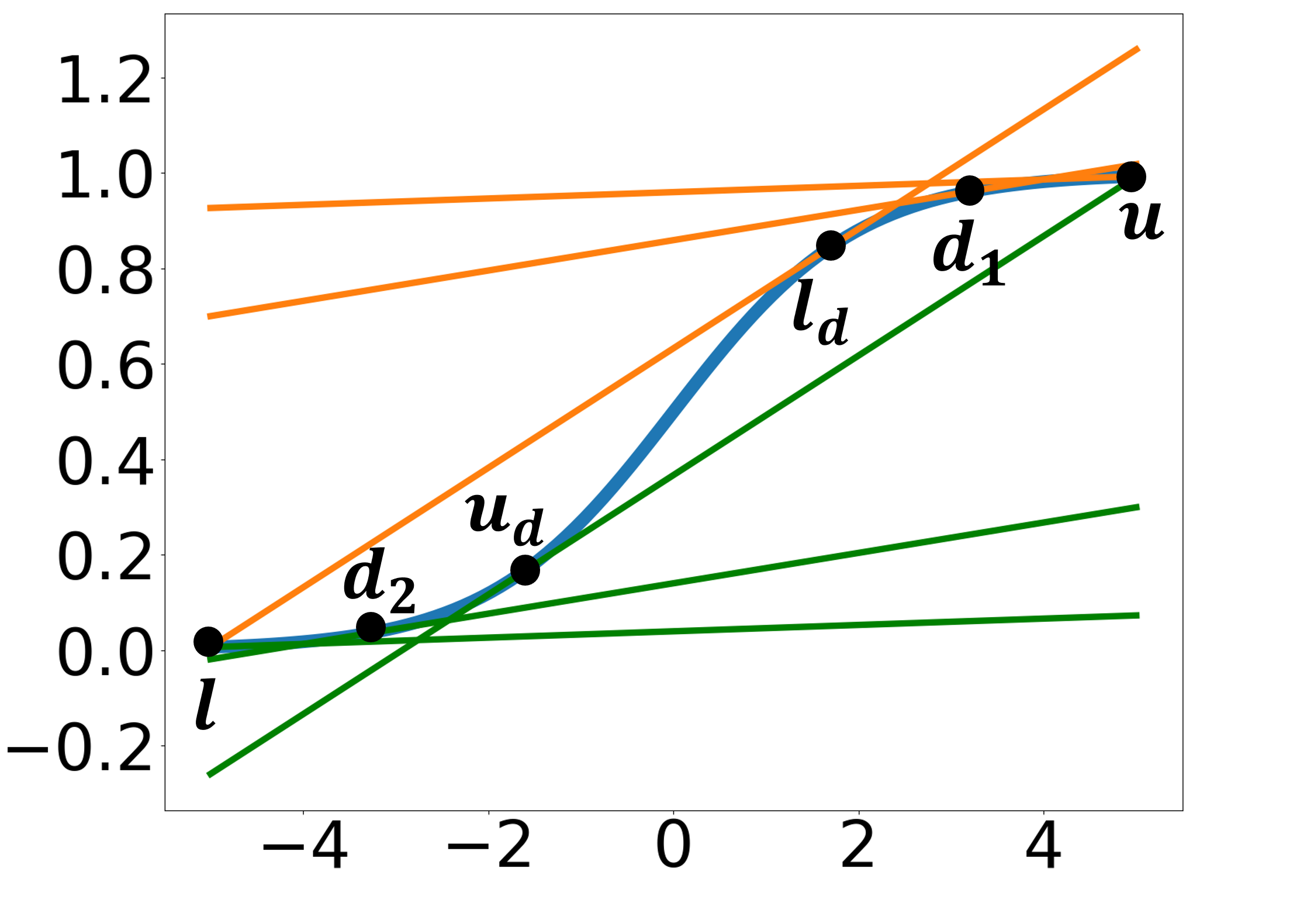}}\\
    \subfigure[$l <  u \leq 0$]{
    \label{fig:sigmoid3}
    \includegraphics[width=0.49\columnwidth]{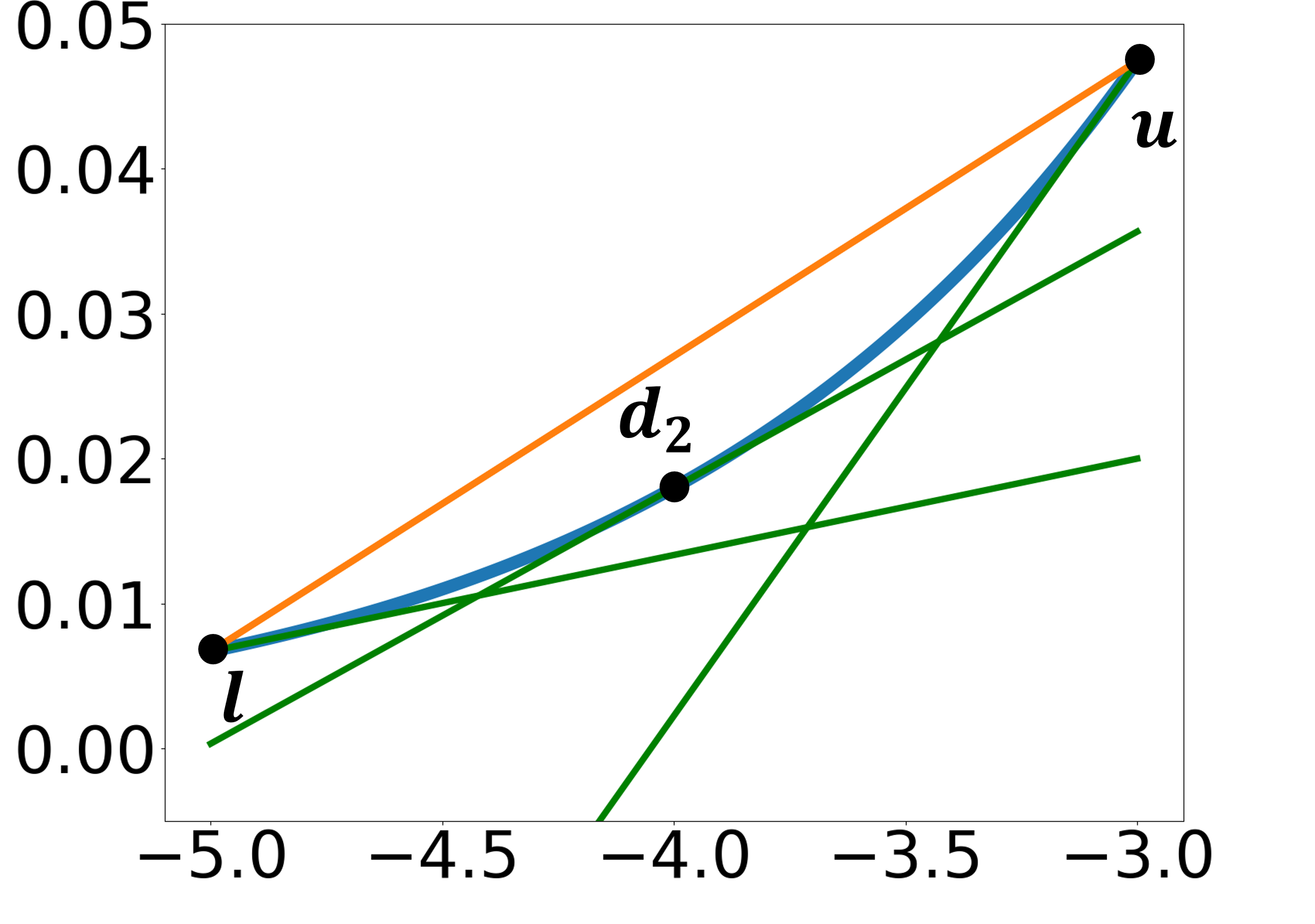}}
    \subfigure[$l < 0 < u$]{
    \label{fig:relu}
    \includegraphics[width=0.49\columnwidth]{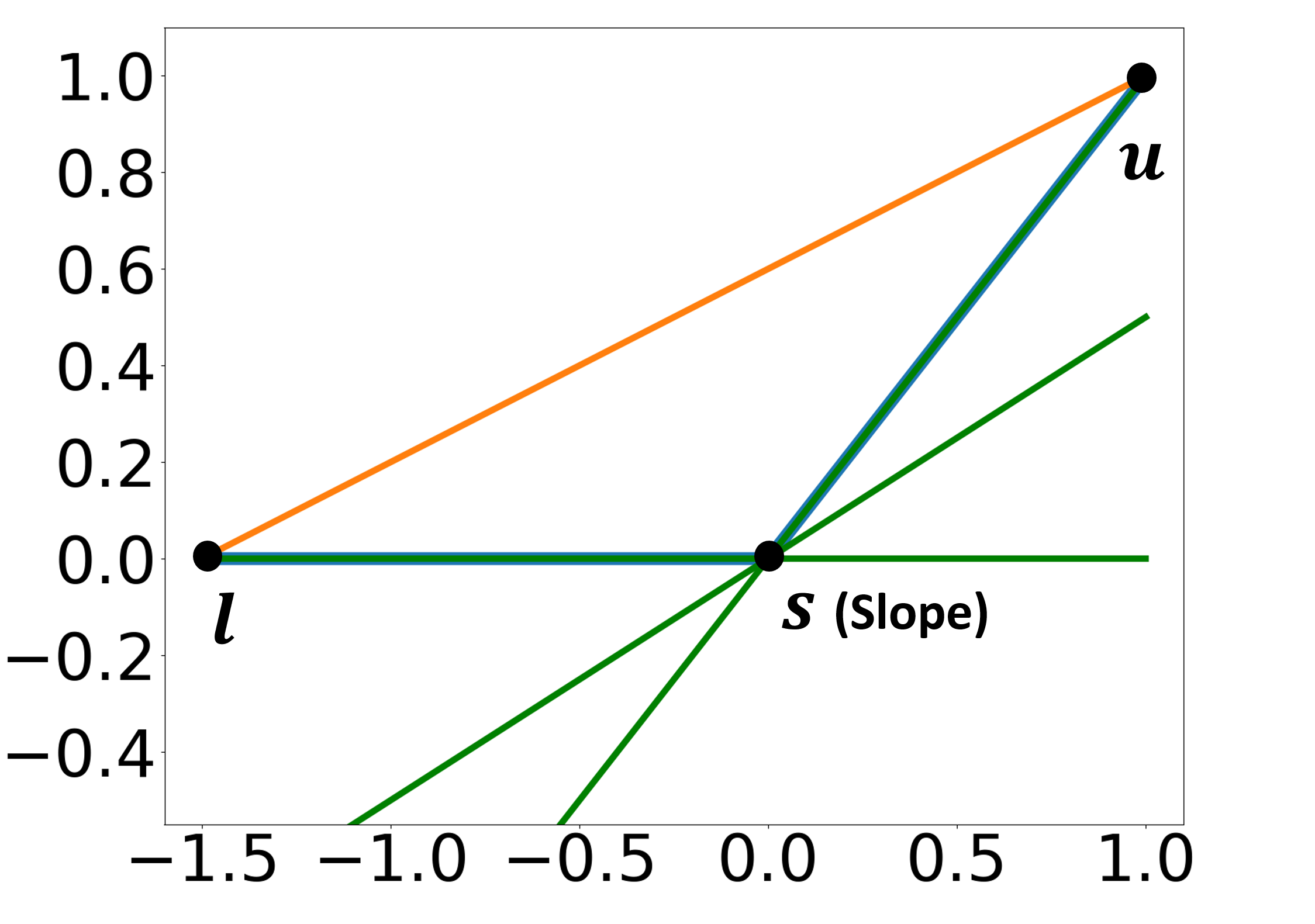}}
    \caption{Illustration of the search space of bounding lines for Sigmoid and ReLU activation. See definitions of $d_1, d_2, u_d, l_d$ in Table~\ref{tbl:bounding_line_search_space}.} 
    \label{fig:sigmoid}
\end{figure}

\section{Fastened CROWN}
\label{sec:fasten}
Recognizing the dependency of lower bounds and upper bounds to slopes and intercepts in the original CROWN method, a consistent use of these parameters is enforced through the self-consistency condition. In fact, we argue that this constraint can be lifted in general (we visualize this relaxation in Figure~\ref{fig:neuronwise_optmize}). The aim of this section stems from this relaxation and focuses on optimizing the pre-activation/output bounds over these tunable bounding parameters to achieve tighter bounds. In that merit, we propose an optimization framework called \textit{FROWN} (\textbf{F}astened C\textbf{ROWN}) for tightening robustness certificates in CROWN. Moreover, FROWN is versatile and can be widely applied to tighten previously-proposed CNN-Cert~\cite{Akhilan2019CNN-Cert} for convolutional neural networks and POPQORN~\cite{ko2019quantifying} for recurrent neural networks. We formalize the objective as the following two optimization problems:
\begin{align}
\label{eqn:max_lb}
&\max_{
s^{[k-1]L},s^{[k-1]U},t^{[k-1]L},t^{[k-1]U}}  \; \gamma_i^{(k)L}\\
\nonumber&\text{s.t.}~  \; s^{(v)L}_i \z^{(v)}_i + t^{(v)L}_i \leq \sigma(\z^{(v)}_i) \leq s^{(v)U}_i \z^{(v)}_i + t^{(v)U}_i,\\
\nonumber&  \!\qquad\forall~ \z^{(v)}_i \in [\mathbf{l}^{(v)}_i, \mathbf{u}^{(v)}_i],  i \in [n_v], v \in [k-1], 
\end{align}
and
\begin{align}
\label{eqn:min_ub}
&\min_{s^{[k-1]L},s^{[k-1]U},t^{[k-1]L},t^{[k-1]U}}  \; \gamma_i^{(k)U}\\
\nonumber&\text{s.t.}~  \; s^{(v)L}_i \z^{(v)}_i + t^{(v)L}_i \leq \sigma(\z^{(v)}_i) \leq s^{(v)U}_i \z^{(v)}_i + t^{(v)U}_i,\\
\nonumber& \!\qquad\forall~ \z^{(v)}_i \in [\mathbf{l}^{(v)}_i, \mathbf{u}^{(v)}_i], i \in [n_v], v \in [k-1].
\end{align}

However, we stress that Problems~\eqref{eqn:max_lb} and~\eqref{eqn:min_ub} are generally non-convex optimization problems when there are more than two layers in the target network. We enclose the proof as Section~A.7 in the appendix. Therefore, optimizing for a non-convex objective function over parameters in large search spaces with infinite number of constraints is impractical. To this end, our ideas are to limit the search space to smaller ones. We present our solutions by first introducing the ideas of ``tighter'' bounding lines.
\begin{definition}
Suppose $\tilde{h}^{(k)L}_i(\z^{(k)}_i)=\tilde{s}^{(k)L}_i\z^{(k)}_i+\tilde{t}^{(k)L}_i$ and $\hat{h}^{(k)L}_i(\z^{(k)}_i)=\hat{s}^{(k)L}_i\z^{(k)}_i+\hat{t}^{(k)L}_i$ are two lower-bounding lines that satisfy

$\begin{cases}
    \tilde{h}^{(k)L}_i(\z^{(k)}_i) > \hat{h}^{(k)L}_i(\z^{(k)}_i), ~\forall~ \z^{(k)}_i \in (\mathbf{l}^{(k)}_i, \mathbf{u}^{(k)}_i)\\
    \sigma(\z^{(k)}_i) \geq \tilde{h}^{(k)L}_i(\z^{(k)}_i), ~\forall~ \z^{(k)}_i \in [\mathbf{l}^{(k)}_i, \mathbf{u}^{(k)}_i]
\end{cases}$
then we say $\tilde{h}^{(k)L}_i(\z^{(k)}_i)=\tilde{s}^{(k)L}_i\z^{(k)}_i+\tilde{t}^{(k)L}_i$ is a tighter lower-bounding line than $\hat{h}^{(k)L}_i(\z^{(k)}_i)=\hat{s}^{(k)L}_i\z^{(k)}_i+\hat{t}^{(k)L}_i$ for the nonlinear activation $\sigma$ in the interval $[\mathbf{l}^{(k)}_i, \mathbf{u}^{(k)}_i]$.
\end{definition}
Similarly, we define a tighter upper-bounding line. Accordingly, the tightest bounding line refers to the $\hat{h}^{(k)L}_i$ when there is, by definition, no tighter bounding line than itself. Note that the tightest bounding line may not be unique. For example, any line passing through the origin with a slope between $0$ and $1$ is a tightest lower-bounding line for the ReLU activation in an interval across the origin ($\mathbf{l}^{(v)}_i<0<\mathbf{u}^{(v)}_i$). With the notion of tightness, a straightforward idea is to adopt one of the tightest bounding lines in every layer for generally tighter closed-form pre-activation/output bounds. However, the proposition:
\begin{quote}
    tighter bounding lines $\Rightarrow$ tighter closed-form bounds
\end{quote}
is not always true. We observe tighter bounding lines can sometimes lead to looser bounds. However, if we roll back to Condition~\ref{cdt:self_consistency}, we can prove that it constitutes a sufficient condition for the proposition.
\begin{theorem}
\label{thm:tighter}
If the robustness of a neural network is evaluated by CROWN on two trials with two different sets of bounding lines characterized by $\{\tilde{s}^{[k-1]U},\tilde{s}^{[k-1]L},\tilde{t}^{[k-1]U},\tilde{t}^{[k-1]L}\}$ and $\{\hat{s}^{[k-1]U},\hat{s}^{[k-1]L},\hat{t}^{[k-1]U},\hat{t}^{[k-1]L}\}$, and in both of which the self-consistency condition is met, then the closed-form bounds obtained via CROWN satisfy
\begin{align*}
&\gamma_i^{(k)L}(\tilde{s}^{[k-1]U},\tilde{s}^{[k-1]L},\tilde{t}^{[k-1]U},\tilde{t}^{[k-1]L}) \geq \\
&\gamma_i^{(k)L}(\hat{s}^{[k-1]U},\hat{s}^{[k-1]L},\hat{t}^{[k-1]U},\hat{t}^{[k-1]L}), \\
&\gamma_i^{(k)U}(\tilde{s}^{[k-1]U},\tilde{s}^{[k-1]L},\tilde{t}^{[k-1]U},\tilde{t}^{[k-1]L}) \leq \\
&\gamma_i^{(k)U}(\hat{s}^{[k-1]U},\hat{s}^{[k-1]L},\hat{t}^{[k-1]U},\hat{t}^{[k-1]L}), 
\end{align*}
for $\forall~ i \in [n_k]$, when bounding lines determined by $\{\tilde{s}^{[k-1]U},\tilde{s}^{[k-1]L},\tilde{t}^{[k-1]U},\tilde{t}^{[k-1]L}\}$ are the same as or tighter than those determined by $\{\hat{s}^{[k-1]U},\hat{s}^{[k-1]L},$ $\hat{t}^{[k-1]U},\hat{t}^{[k-1]L}\}$.
\end{theorem}
Proof:
The self-consistency guarantees the optimality of bounds given by CROWN to the corresponding LP problem~\eqref{eqn:LP} according to Theorem~\ref{thm:crownoptimal}.
As the use of tighter bounding lines in Problem~\eqref{eqn:LP} narrows the feasible set, the optimal value of Problem~\eqref{eqn:LP} (lower pre-activation/output bound) will stay or grow larger, which means the lower bound given by CROWN will stay or grow larger. Similar arguments extend to tightened upper bounds.\qed

Till now, we have confirmed the connection between the tightest bounding lines and the tightest CROWN pre-activation/output bound under Condition~\ref{cdt:self_consistency}. 
In addition, we manage to prove Theorem~\ref{thm:tighter} under a weaker condition (see its proof in~Section~A.6 in the appendix).

Condition~\ref{cdt:self_consistency} is too strong a condition for our proposed optimization framework to be practical. Actually we propose to improve CROWN by breaking Condition~\ref{cdt:self_consistency}.
Our problem can be eased by only considering the dependency of closed-form pre-activation/output bounds on the intercepts (with the slopes fixed).
We provide the following theorem:
\begin{theorem}
\label{thm:tighter_slopes}
If the robustness of a neural network is evaluated by CROWN on two trials with bounding lines characterized by $\{s^{[k-1]U},s^{[k-1]L},\tilde{t}^{[k-1]U},\tilde{t}^{[k-1]L}\}$ and $\{s^{[k-1]U},s^{[k-1]L},\hat{t}^{[k-1]U},\hat{t}^{[k-1]L}\}$, then the closed-form bounds obtained via CROWN satisfy
\begin{align*}
&\gamma_i^{(k)L}({s}^{[k-1]U},{s}^{[k-1]L},\tilde{t}^{[k-1]U},\tilde{t}^{[k-1]L}) \geq \\
&\gamma_i^{(k)L}({s}^{[k-1]U},{s}^{[k-1]L},\hat{t}^{[k-1]U},\hat{t}^{[k-1]L}), \\
&\gamma_i^{(k)U}({s}^{[k-1]U},{s}^{[k-1]L},\tilde{t}^{[k-1]U},\tilde{t}^{[k-1]L}) \leq \\
&\gamma_i^{(k)U}({s}^{[k-1]U},{s}^{[k-1]L},\hat{t}^{[k-1]U},\hat{t}^{[k-1]L}),
\end{align*}
for $\forall~ i \in [n_k]$, when $\tilde{t}^{(v-1)L} \succcurlyeq \hat{t}^{(v-1)L}$ and $\tilde{t}^{(v-1)U} \preccurlyeq \hat{t}^{(v-1)U}$, $\forall v \in [k-1]$.
\end{theorem}
This theoretical guarantee limits the freedom in choosing the intercepts: we should always choose upper-bounding lines with smaller intercepts and lower-bounding lines with larger intercepts if different bounding lines are allowed for different network neurons. 
Note that this conclusion holds under no assumptions on the choice of bounding lines and hence can be used to instruct how to choose bounding lines in FROWN.
We demonstrate Theorem~\ref{thm:tighter_slopes} can be used to reduce the search space of the upper (or lower) bounding line to one that can be characterized by a single variable continuously in Appendix Section A.8. This enables the usage of gradient-based method to search over candidate bounding lines to obtain tighter bounds.
We further limit the search space to the tightest bounding lines (which is a subset of the search space narrowed only by Theorem~\ref{thm:tighter_slopes}) as demonstrated in Table~\ref{tbl:bounding_line_search_space} and exemplified in Figure~\ref{fig:sigmoid} in order to simplify implementation.
We emphasize that this limit is not necessary. FROWN can be readily generalized to the case in which the search space is reduced only by Theorem~\ref{thm:tighter_slopes}, and the obtained bounds should be even tighter as the search space is larger.
Since the tightest bounding lines defined in Table~\ref{tbl:bounding_line_search_space} automatically satisfy the optimization constraints in Problems~\eqref{eqn:max_lb} and~\eqref{eqn:min_ub}, the constrained optimization problems are then converted to unconstrained ones. 
Furthermore, notice that the objective functions in the two problems are differentiable to bounding line parameters. 
This allows us to solve the problems by projected gradient descent~\cite{nesterov2014convex} (see details in Appendix Section A.9). 

By and large, given an $m$-layer network $F$, input sample $\x_0\in\mathbb{R}^{n}$, $l_p$ ball parameters $p\geq 1$, and $\epsilon\ge 0$, for $\forall~j\in[n_m]$, $1/q=1-1/p$, we can compute two fixed values $\gamma^L_j$ and $\gamma^U_j$ such that $\forall \x\in\mathbb{B}_p(\x_0,\epsilon)$, the inequality $\gamma^L_j\leq F_j(\x)\leq \gamma^U_j$ holds. 
Suppose the label of the input sequence is $i$, the largest possible lower bound $\epsilon_i$ of untargeted and targeted (target class $j$) attacks is found by solving:
\begin{align*}
    \text{Untargeted:}\qquad\epsilon_i &=\max_{\epsilon}\epsilon,\;
    \textrm{\textrm{s.t.}}\; \gamma^L_i(\epsilon ) \geq \gamma^U_j(\epsilon) ,\; \forall j\neq i.\\
    \text{Targeted:}\quad\hat \epsilon(i,j) &=\max_{\epsilon}\epsilon,\;
    \textrm{\textrm{s.t.}}\; \gamma^L_i(\epsilon ) \geq \gamma^U_j(\epsilon).
\end{align*}
We conduct binary search procedures to compute the largest possible $\epsilon_{i}$ (or $\hat \epsilon$).

\begin{table*}[t!]
\centering
\caption{(Experiment I) Averaged certified $l_{\infty}$ bounds and $l_p$ bounds ($p=1,2,\infty$) of Sensorless Drive Diagnosis classifiers and MNIST classifiers, respectively. "N/A" indicates no results can be obtained in the given runtime. The up arrow ``$\uparrow$'' means `` more than''.
``$m \times [N] \; \sigma$'' means an $m$-layer network with $N$ neurons and $\sigma$ activation.
}
\label{tbl:exp1}
\resizebox{1.9\columnwidth}{!}{
\npdecimalsign{.}
\nprounddigits{4}
\npthousandsep{}
\begin{tabular}{c|c|n{2}{4} n{2}{4} n{2}{4} | >{{\nprounddigits{2}}}n{2}{2} >{{\nprounddigits{2}}}n{2}{2} |>{{\nprounddigits{2}}}n{3}{2} >{{\nprounddigits{2}}}n{4}{2} | >{{\nprounddigits{1}}}n{6}{1}}
    \hline
    \multicolumn{10}{c}{Sensorless Drive Diagnosis classifiers}\\\hline
    \multirow{2}{*}{Network}    & \multirow{2}{*}{p} & \multicolumn{3}{c|}{Certified Bounds} & \multicolumn{2}{c|}{Improvement} & \multicolumn{2}{c|}{Avg. Time per Image (s)} & {\quad Speedups of} \\
    \cline{3-9}
    &                    & CROWN & FROWN & {\quad \;LP}        & FROWN        & {\quad LP}         & FROWN    & {\quad \;\;LP}                 & {FROWN over LP} \\
    \hline
    $4 \times [20]$ ReLU & \multirow{3}{*}{$\infty$}    & 0.20189309 & 0.22465439 & 0.22689147 & 11.2739\% & \textbf{12.38\%} & 0.346491119 & 0.960072917 & 2.7708{\,X} \\
    $8 \times [20]$ ReLU &    & 0.20936128 & 0.23646373 & 0.25261395 & 12.9453\% & \textbf{20.66\%} & 1.808810924 & 4.194196891 & 2.3188{\,X} \\
    $12 \times [20]$ ReLU &   & 0.19960524 & 0.24960147 & 0.27402096 & 25.0476\% & \textbf{37.28\%} & 4.388828654 & 9.455646552 & 2.1545{\,X} \\
    \hline
    $4 \times [20]$ Sigmoid &\multirow{3}{*}{$\infty$}  & 0.10192142 & 0.14175183 & 0.13882568 & \textbf{39.08\%} & 36.2086\% & 0.743399554 & 2.114387755 & 2.8442{\,X} \\
    $8 \times [20]$ Sigmoid  & & 0.08578154 & 0.16179805 & 0.16263148 & 88.6164\% & \textbf{89.59\%} & 4.149629630 & 32.14590909 & 7.7467{\,X} \\
    $12 \times [20]$ Sigmoid & & 0.07820402 & 0.15097938 & 0.10809507 & \textbf{93.06\%} & 38.2219\% & 8.710052219 & 152.9056907 & 17.555{\,X} \\
    \hline
    \multicolumn{10}{c}{MNIST classifiers}\\
    \hline
    \multirow{3}{*}{\makecell{$5 \times [20]$ \\ ReLU}} & 1        & 3.80175090 & 4.08353186 & 4.22146046 & 7.4119\%  & \textbf{11.04\%} & 1.689364 & 195.0664 & 115.4674{\,X} \\
    & 2        & 0.53455848 & 0.57095152 & 0.55870703 & \textbf{\; 6.81\%}  & 4.52\%  & 1.122871 & 41.6283  & 37.0731{\,X}  \\ 
    & $\infty$ & 0.02610023 & 0.02780137 & 0.02868102 & 6.5177\%  & \textbf{\; 9.89\%}  & 1.258348 & 11.93404 & 9.4839{\,X} \\ 
    \hline
    \multirow{3}{*}{\makecell{$20 \times [20]$ \\ ReLU}}& 1        & 2.3853     & 3.1062     & 3.17351974 & 30.2226\% & \textbf{33.04\%} & 35.5283  & 1301.114 & 36.6219{\,X} \\
    & 2        & 0.3656     & 0.4925     & 0.50737058 & 34.7101\% & \textbf{38.78\%} & 31.10188 & 229.8665 & 7.3908{\,X} \\ 
    & $\infty$ & 0.01834928 & 0.02400810 & 0.02454141 & 30.8395\% & \textbf{33.75\%} & 43.31122 & 199.4015 & 4.6039{\,X} \\ 
    \hline
    \multirow{3}{*}{\makecell{$5 \times [20]$ \\ Sigmoid}} & 1        & 1.80092955 & 2.13400507 & 2.11258952 & \textbf{18.49\%} & 17.3055\% & 1.751429 & 310.2723 & 177.1538{\,X} \\ 
    & 2        & 0.31002384 & 0.35811698 & 0.34165625 & \textbf{15.51\%} & 10.2032\% & 1.226073 & 44.0025  & 35.8890{\,X} \\ 
    & $\infty$ & 0.01530928 & 0.01744410 & 0.01700973 & \textbf{13.94\%} & 11.1073\% & 1.385459 & 17.19214 & 12.4090{\,X} \\ 
    \hline
    \multirow{3}{*}{\makecell{$20 \times [20]$ \\ Sigmoid}}& 1        & 1.53479946 & 1.97794294 & 1.97303025 & \textbf{28.87\%} & 28.5530\% & 31.79776 & 4904.991 & 154.2559{\,X} \\ 
    & 2        & 0.25236204 & 0.32607010 & 0.26567794 & \textbf{29.21\%} & 5.2765 \% & 31.76949 & 1354.536 & 42.6364{\,X} \\ 
    & $\infty$ & 0.01308495 & 0.01661916 & 0.01532545 & \textbf{27.01\%} & 17.1227\% & 44.86479 & 1859.683 & 41.4508{\,X} \\ 
    \hline
    \multirow{3}{*}{\makecell{$5 \times [100]$ \\ ReLU}} & 1        & 3.8928     & 4.23426247 &{\quad N/A} & \textbf{\; 8.77\%}  &{\quad N/A}& 13.10667 & 6000 {$\uparrow$}   &457.7822{\,X} {$\uparrow$}\\ 
    & 2        & 0.5499     & 0.57894945 & 0.59337910 & 5.2827\%  & \textbf{\,\,7.91\%}  & 10.89919 & 826.7916 & 75.8581{\,X}  \\
    & $\infty$ & 0.02605076 & 0.02757367 & 0.02776357 & 5.8459\%  & \textbf{\,\,6.57\%}  & 10.41608 & 446.0726 & 42.8254{\,X}  \\ 
    \hline
    \multirow{3}{*}{\makecell{$7 \times [100]$ \\ ReLU}} & 1        & 3.5509     & 3.99067092 &{\quad N/A} & \textbf{12.38\%} &{\quad N/A}& 55.07379 & 10000 {$\uparrow$}  &181.5746{\,X} {$\uparrow$}\\ 
    & 2        & 0.4997     & 0.53474498 &{\quad N/A} & \textbf{\; 7.01\%}  &{\quad N/A}& 49.48793 & 10000 {$\uparrow$}  &202.0695{\,X} {$\uparrow$}\\ 
    & $\infty$ & 0.02411035 & 0.02582039 &{\quad N/A} & \textbf{\; 7.09\%}  &{\quad N/A}& 44.93692 & 10000 {$\uparrow$}  &222.5342{\,X} {$\uparrow$}\\ 
    \hline
    \multirow{3}{*}{\makecell{$5 \times [100]$ \\ Sigmoid}} & 1        & 2.09984541 & 2.51844215 &{\quad N/A} & \textbf{19.93}\% &{\quad N/A}& 13.64317 & 10000 {$\uparrow$}    &732.9675{\,X}{$\uparrow$} \\ 
    & 2        & 0.32327360 & 0.37738118 & 0.24942495 & \textbf{16.74\%} & -22.8440\%& 12.37509 & 9368.58454 & 757.0516{\,X} \\ 
    & $\infty$ & 0.01562966 & 0.01856160 & 0.01767777 & \textbf{18.76\%} & 13.1040\% & 13.09035 & 1319.25163 & 100.7805{\,X} \\ 
    \hline
    \multirow{3}{*}{\makecell{$7 \times [100]$ \\ Sigmoid}} & 1        & 1.72942972 & 2.23796749 &{\quad N/A} & \textbf{29.40\%} &{\quad N/A}& 48.95841 & 10000 {$\uparrow$}    &204.2550{\,X} {$\uparrow$}\\                       & 2        & 0.28328878 & 0.34267208 &{\quad N/A} & \textbf{20.96\%} &{\quad N/A}& 42.0407  & 10000 {$\uparrow$}    &237.8647{\,X} {$\uparrow$}\\ 
    & $\infty$ & 0.01399797 & 0.01679174 &{\quad N/A} & \textbf{19.96\%} &{\quad N/A}& 41.99858 & 10000 {$\uparrow$}    &238.1033{\,X} {$\uparrow$}\\ 
    \hline
\end{tabular}
}

\centering
\caption{(Experiment II) Averaged certified $l_p$ bounds of different classifiers on CIFAR10 Networks.}
\resizebox{2.1\columnwidth}{!}{
\npdecimalsign{.}
\nprounddigits{4}
\npthousandsep{}
\begin{tabular}{c|c|n{2}{4} n{2}{4}| >{{\nprounddigits{2}}}n{2}{2} |c|c|n{2}{4} n{2}{4} |>{{\nprounddigits{2}}}n{2}{2} |c|c|n{2}{4} n{2}{4} |>{{\nprounddigits{2}}}n{2}{2}}
    \hline
    \multirow{2}{*}{Network} & \multirow{2}{*}{p} & \multicolumn{2}{c|}{Certified Bounds} & {Improve-}   & \multirow{2}{*}{Network} & \multirow{2}{*}{p} & \multicolumn{2}{c|}{Certified Bounds} & {Improve-} & \multirow{2}{*}{Network} & \multirow{2}{*}{p} & \multicolumn{2}{c|}{Certified Bounds} & {Improve-}\\
    \cline{3-4} \cline{8-9}
    &                    & CROWN & FROWN                         &  {\;\; ment} &                          &                    & CROWN & FROWN                         & {\;\;ment} &                          &                    & CROWN & FROWN                         & {\;\;ment}  \\
    \hline
    \multirow{3}{*}{\makecell{$4 \times [2048]$ \\ ReLU}}   & 1     & 7.54595089 & 7.59892654 & 0.7020\% & \multirow{3}{*}{\makecell{$6 \times [2048]$ \\ ReLU}}   & 1        & 4.59904099 & 4.72408152 & 2.7188\%  & \multirow{3}{*}{\makecell{$8 \times [2048]$ \\ ReLU}}   & 1        & 4.23493385 & 4.94163609 &16.6874\% \\
    & 2 & 0.61745107 & 0.63034403 & 2.0881\% & & 2 & 0.37454715 & 0.37229005 &-0.6026\%  & & 2 & 0.34760612 & 0.38093299 & 9.5875\% \\
    & $\infty$ & 0.01506771 & 0.01569192 & 4.1427\%  & & $\infty$ & 0.00918379 & 0.00927143 & 0.9543\%  & & $\infty$ & 0.00866792 & 0.00943343 & 8.8315\% \\\hline
    \multirow{3}{*}{\makecell{$4 \times [2048]$ \\ Sigmoid}} & 1    & 3.65584946 & 4.47255278 & 22.3396\% & \multirow{3}{*}{\makecell{$6 \times [2048]$ \\ Sigmoid}} & 1        & 1.50934732 & 1.84298861 & 22.1050\% & \multirow{3}{*}{\makecell{$8 \times [2048]$ \\ Sigmoid}} & 1        & 1.28750134 & 1.68622041 & 30.9684\% \\ 
    & 2 & 0.28469625 & 0.33529428 & 17.7726\% & & 2 & 0.11799628 & 0.14479597 & 22.7123\% & & 2 & 0.10062468 & 0.12945828 & 28.6546\%\\
    & $\infty$ & 0.00690464 & 0.00817044 & 18.3326\% & & $\infty$ & 0.00285764 & 0.00345873 & 21.0345\%  &  & $\infty$ & 0.00237977 & 0.00327576 & 37.6503\% \\
    \hline
\end{tabular}
}
\end{table*}

\section{Experimental Results}
\label{sec:exp}
\paragraph{Overview.}
In this section, we aim at comparing the LP-based method\footnote{The highly-efficient Gurobi LP solver is adopted here.} and FROWN as two approaches to improve CROWN. 
We allow the LP-based method to use more than one bounding lines (which also increases computation cost) in order to make improvement on CROWN.
Specifically, two lower-bounding lines are considered for ReLU networks while up to three upper/lower-bounding lines are adopted for Sigmoid (or Tanh) networks in the LP-based method (more details supplemented as Section A.4 in the appendix).
On the other hand, FROWN improves CROWN solutions by optimizing over the bounding lines to give tighter bounds. 
These two approaches are evaluated and compared herein by both the safeguarded regions they certify and their time complexity. 
We run the LP-based method on a single Intel Xeon E5-2640 v3 (2.60GHz) CPU. 
We implement our proposed method FROWN using PyTorch to enable the use of an NVIDIA GeForce GTX TITAN X GPU. 
However, we time FROWN on a single Intel Xeon E5-2640 v3 (2.60GHz) CPU when comparing with the LP-based method for fair comparisons. 
We leave the detailed experimental set-ups and complete experimental results to Appendix Section~A.9.
\paragraph{Experiment I. }
In the first experiment, we compare the improvements of FROWN and the LP-based method over CROWN on sensorless drive diagnosis networks~\footnote{\url{https://archive.ics.uci.edu/ml/datasets/Dataset+for+Sensorless+Drive+Diagnosis}} and MNIST classifiers. We present their results in Table~\ref{tbl:exp1}. As shown in the table, we consider ReLU and Sigmoid (results of Tanh networks are included in Appendix Section~A.9) networks that are trained independently on two datasets. 
The size of the networks ranges from $3$ layers to $20$ layers and $20$ neurons to $100$ neurons. 
We remark that even on networks with only $100$ neurons, the LP-based method scales badly and is unable to provide results in $100$ minutes for only one image. 
The improved bounds in Table~\ref{tbl:exp1} verify the effectiveness of both FROWN and LP-based approach in tightening CROWN results. Specifically, an up to $93\%$ improvement in the magnitude of bounds is witnessed on sensorless drive diagnosis networks. 
And in general, the deeper the target network is, the greater improvement can be achieved. 
When comparing FROWN to the LP-based method, it is demonstrated that FROWN computes bounds up to two orders of magnitudes faster than the LP-based method and is especially advantageous when certifying $l_1$-norm regions. 
On the other hand, while the LP-based method gives larger certified bounds for ReLU networks in most cases, FROWN certifies larger bounds for Sigmoid and Tanh networks.

\paragraph{Experiment II.}
In our second experiment, we compute the robustness certificate on CIFAR10 networks that have $2048$ neurons in each layer. With the width of neural networks, the LP-based method is \textbf{unusable} due to its high computational-complexity. Therefore, we only show the improvements FROWN has brought to the original CROWN solutions. 
In this experiment, we further speed up FROWN by optimizing neurons in a layer group by group, instead of one by one, and we provide a parameter to balance the trade-off between tightness of bounds and time cost in FROWN(see details in Appendix Section A.10).
We observe consistent trends on CIFAR10 networks: the deeper the neural network is, the more significant improvement can be made by FROWN in tightening CROWN solutions. Notably, an up to $38\%$ improvement in certified bounds is achieved when considering $l_{\infty}$-norm balls in Sigmoid networks.

\subsection{Discussion}
Overall, we have shown the trade-off between the computational costs and certified adversarial distortions in ReLU networks:  
the LP-based approach certifies larger bounds than FROWN at the cost of $2$ to $178$ times longer runtime. However, the LP-based approach suffers poor scalability and soon becomes computationally infeasible as the network grows deeper or wider. In contrast, FROWN manages to increase the certified region of CROWN much more efficiently and wins over the LP-based approach in almost all Sigmoid/Tanh networks.
Notably, in some cases, the LP-based method gives even worse result than CROWN (those with negative improvements). We conclude two possible reasons: i) the Gurobi LP solver is not guaranteed to converge to the optimal solution and ii) statistical fluctuations caused by random sample selections. More discussions on this are included in Appendix Section~A.9.

\section{Conclusion}
\label{sec:conclude}
In this paper, we have proved the optimality of CROWN in the relaxed LP framework under mild conditions. Furthermore, we have proposed a general and versatile optimization framework named \textit{FROWN} for optimizing state-of-the-art formal robustness verifiers including CROWN, CNN-Cert, and POPQORN. Experiments on various networks have verified the usefulness of FROWN in providing tightened robustness certificates at a significantly lower cost than the LP-based method.

\section{Acknowledgement}
This work is partially supported by the General Research Fund (Project 14236516) of the Hong Kong Research Grants Council, and MIT-Quest program.

\bibliographystyle{aaai}
\bibliography{AAAI-LyuZ.3007}

\end{document}